\documentclass[10pt,twocolumn,letterpaper]{article}

\usepackage{cvpr}
\usepackage{times}
\usepackage{epsfig}
\usepackage{graphicx}
\usepackage{amsmath}
\usepackage{amssymb}
\usepackage[dvipsnames]{xcolor}
\usepackage{subcaption}
\usepackage{booktabs}

\usepackage[titletoc,title]{appendix}
\usepackage[abs]{overpic}

\usepackage{tabulary,multirow}

\newcolumntype{x}[1]{>{\centering\arraybackslash}p{#1pt}}
\newlength\savewidth

\usepackage[pagebackref=true,breaklinks=true,letterpaper=true,colorlinks,bookmarks=false]{hyperref}

\cvprfinalcopy 

\def\cvprPaperID{3583} 

\ifcvprfinal\fi
\begin{document}

\title{NAS-FPN: Learning Scalable Feature Pyramid Architecture\\ for Object Detection}

\author{Golnaz Ghaisi \quad Tsung-Yi Lin \quad Ruoming Pang \quad Quoc V. Le\\
Google Brain\\
{\tt\small \{golnazg,tsungyi,rpang,qvl\}@google.com}
}

\maketitle
\thispagestyle{empty}

\begin{abstract}
Current state-of-the-art convolutional architectures for object detection are manually designed. Here we aim to learn a better architecture of feature pyramid network for object detection. We adopt Neural Architecture Search and discover a new feature pyramid architecture in a novel scalable search space covering all cross-scale connections. The discovered architecture, named NAS-FPN, consists of a combination of top-down and bottom-up connections to fuse features across scales. NAS-FPN, combined with various backbone models in the RetinaNet framework, achieves better accuracy and latency tradeoff compared to state-of-the-art object detection models. NAS-FPN improves mobile detection accuracy by 2 AP compared to state-of-the-art SSDLite with MobileNetV2 model in~\cite{sandler2018mobilenetv2} and achieves 48.3 AP which surpasses Mask R-CNN~\cite{Detectron2018} detection accuracy with less computation time.
\end{abstract}

\section{Introduction}
Learning visual feature representations is a fundamental problem in computer vision. In the past few years, great progress has been made on designing the model architecture of deep convolutional networks (ConvNets) for image classification \cite{he2016resnet,huang2017densenet,szegedy2015googlenet} and object detection \cite{li2018detnet,lin2017fpn}. Unlike image classification which predicts class probability for an image, object detection has its own challenge to detect and localize multiple objects across a wide range of scales and locations. To address this issue, the pyramidal feature representations, which represent an image with multiscale feature layers, are commonly used by many modern object detectors \cite{he2017mask,lin2018focal,liu2016ssd}.

\begin{figure}[t]
\centering
\begin{overpic}[width=.99\linewidth]{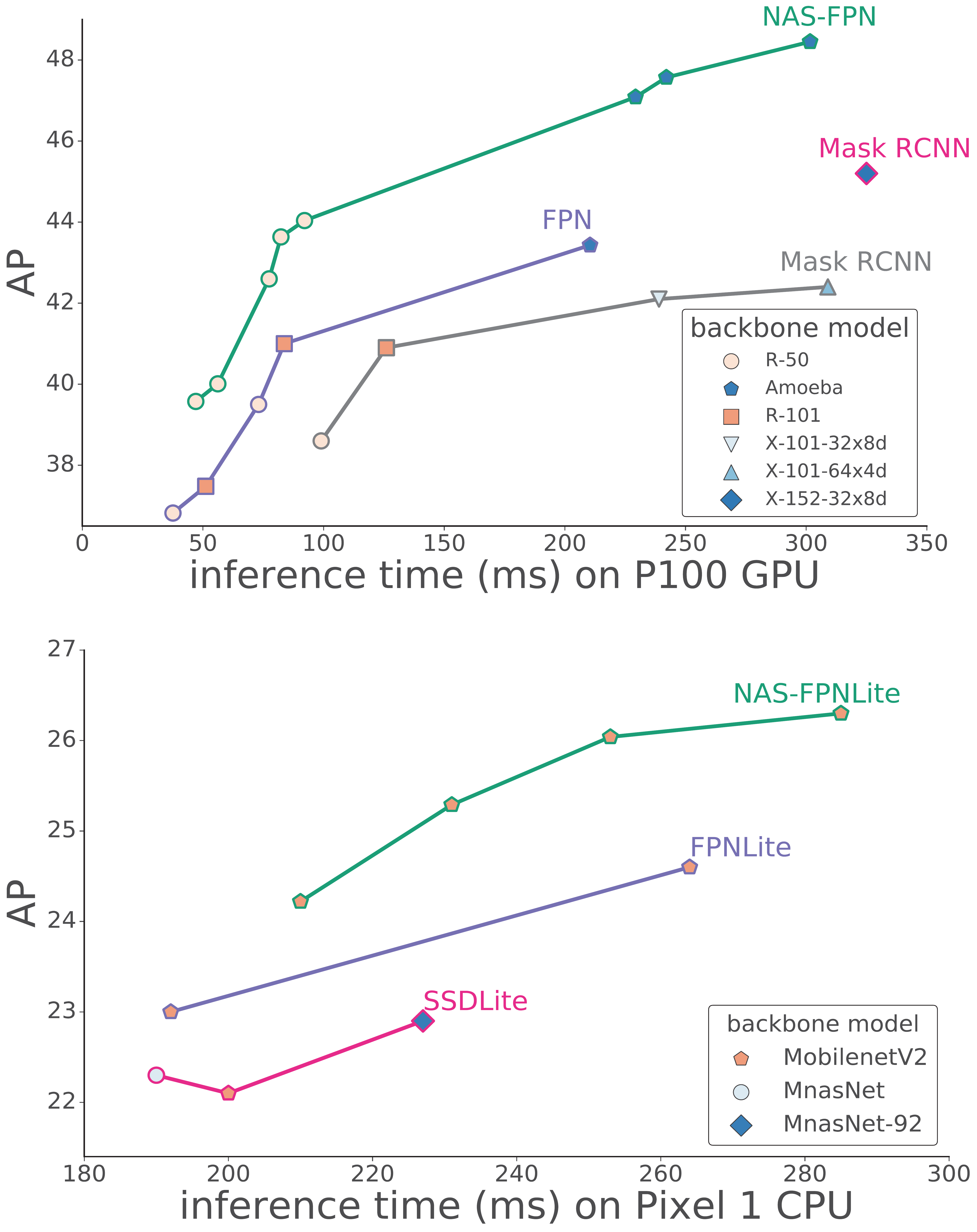}
 \put(225,231){\footnotesize{\cite{Detectron2018}}}
 \put(144,241){\footnotesize{\cite{lin2018focal}}}
 \put(128,53){\footnotesize{\cite{sandler2018mobilenetv2}}}
\end{overpic}
\caption{Average Precision vs. inference time per image across accurate models (top) and fast models (bottom) on mobile device. The green curve highlights results of NAS-FPN combined with RetinaNet. Please refer to Figure~\ref{fig:performance} for details.}
\label{fig:summary}
\end{figure}

Feature Pyramid Network (FPN) \cite{lin2017fpn} is one of the representative model architectures to generate pyramidal feature representations for object detection. It adopts a backbone model, typically designed for image classification, and builds feature pyramid by sequentially combining two adjacent layers in feature hierarchy in backbone model with top-down and lateral connections. The high-level features, which are semantically strong but lower resolution, are upsampled and combined with higher resolution features to generate feature representations that are both high resolution and semantically strong. Although FPN is simple and effective, it may not be the optimal architecture design. Recently, PANet \cite{liu2018panet} shows adding an extra bottom-up pathway on FPN features improves feature representations for lower resolution features. Many recent works \cite{fu2017dssd,liu2018reconfigfpn,kong2016ron,kim2018pfpnet,woo2018stairnet,kim2018san,yu2018dln,xu2018stdn,zhang2018refinedet} propose various cross-scale connections or operations to combine features to generate pyramidal feature representations.

The challenge of designing feature pyramid architecture is in its huge design space. The number of possible connections to combine features from different scales grow exponentially with the number of layers. Recently, Neural Architecture Search algorithm \cite{zoph2017nas} demonstrates promising results on efficiently discovering top-performing architectures for image classification in a huge search space. To achieve their results, Zoph et al.~\cite{zoph2017learning} propose a modularized architecture that can be repeated and stacked into a scalable architecture. Inspired by \cite{zoph2017learning}, we propose the search space of scalable architecture that generates pyramidal representations. The key contribution of our work is in designing the search space that covers all possible \textit{cross-scale} connections to generate multiscale feature representations. During the search, we aims to discover an atomic architecture that has identical input and output feature levels and can be applied repeatedly. The modular search space makes searching pyramidal architectures manageable. Another benefit of modular pyramidal architecture is the ability for anytime object detection (or ``early exit''). Although such early exit approach has been attempted~\cite{huang2017multi}, manually designing such architecture with this constraint in mind is quite difficult.

The discovered architecture, named NAS-FPN, offers great flexibility in building object detection architecture. NAS-FPN works well with various backbone model, such as MobileNet~\cite{sandler2018mobilenetv2}, ResNet~\cite{he2016resnet}, and AmoebaNet~\cite{real2018regularized}. It offers better tradeoff of speed and accuracy for both fast mobile model and accurate model. Combined with MobileNetV2 backbone in RetinaNet framework, it outperforms state-of-the-art mobile detection model of SSDLite with MobilenetV2 ~\cite{sandler2018mobilenetv2} by 2 AP given the same inference time. With strong AmoebaNet-D backbone model, NAS-FPN achieves 48.3 AP single model accuracy with single testing scale. The detection accuracy surpasses Mask R-CNN reported in~\cite{Detectron2018} with even less inference time. A summary of our results is shown in Figure~\ref{fig:summary}.

\section{Related Works}
\subsection{Architecture for Pyramidal Representations}
Feature pyramid representations are the basis of solutions for many computer vision applications required multiscale processing \cite{adelson1984pyramid}. However, using Deep ConvNets to generate pyramidal representations by featurizing image pyramid imposes large computation burden. To address this issue, recent works on human pose estimation, image segmentation, and object detection \cite{ghiasi2016pyramid,he2017mask,lin2017fpn,newell2016hourglass,ronneberger2015unet} introduce cross-scale connections in ConvNets that connect internal feature layers in different scales. Such connections effectively enhance feature representations such that they are not only semantically strong but also contain high resolution information. Many works have studied how to improve mutliscale feature presentations. Liu et.al~\cite{liu2018panet} propose an additional bottom-up pathway based on FPN~\cite{lin2017fpn}. Recently, Zhao et al.~\cite{zhao2019m2det} extends the idea to build stronger feature pyramid representations by employing multiple U-shape modules after a backbone model. Kong et al.~\cite{liu2018reconfigfpn} first combine features at all scales and generate features at each scale by a global attention operation on the combined features. Despite it is an active research area, most architecture designs of cross-scale connections remain shallow compared to the backbone model. In addition to manually design the cross-scale connections, \cite{rubio2018shortcut,islam2017gatednetwork} propose to learn the connections through gating mechanism for visual counting and dense label predictions.

In our work, instead of manually designing architectures for pyramidal representations, we use a combination of scalable search space and Neural Architecture Search algorithm to overcome the large search space of pyramidal architectures. We constrain the search to find an architecture that can be applied repeatedly. The architecture can therefore be used for anytime object detection (or ``early exit"). Such early exit idea is related to~\cite{bolukbasi2017adaptive,teerapittayanon2016branchynet}, especially in image classification~\cite{huang2017multi}.

\subsection{Neural Architecture Search}
Our work is closely related to the work on Neural Architecture Search~\cite{zoph2017nas,baker2017designing,zoph2017learning,real2018regularized}. Most notably, Zoph et al.~\cite{zoph2017learning} use a reinforcement learning with a controller RNN to design a cell (or a layer) to obtain a network, called NASNet which achieves state-of-the-art accuracy on ImageNet. The efficiency of the search process is further improved by~\cite{liu2017progressive} to design a network called PNASNet, with similar accuracy to NASNet. Similarly, an evolution method~\cite{real2018regularized} has also been used to design AmoebaNets that improve upon NASNet and PNASNet. Since reinforcement learning and evolution controllers perform similarly well, we only experiment with a Reinforcement Learning controller in this paper. Our method has two major differences compared to~\cite{zoph2017nas}: (1) the outputs of our method are multiscale features whereas output of \cite{zoph2017nas} is single scale features for classification; (2) our method specifically searches cross-scale connections, while \cite{zoph2017nas} only focuses on discovering connections within the same feature resolution.
Beyond image classification, Neural Architecture Search has also been used to improve image segmentation networks~\cite{chen2018searching}. To the best of our knowledge, our work is the first to report success of applying Neural Architecture Search for pyramidal architecture in object detection. For a broader overview of related methods for Neural Architecture Search, please see~\cite{elsken2018neural}.

\section{Method}
\begin{figure}[h!]
\centering
\includegraphics[width=0.4\textwidth]{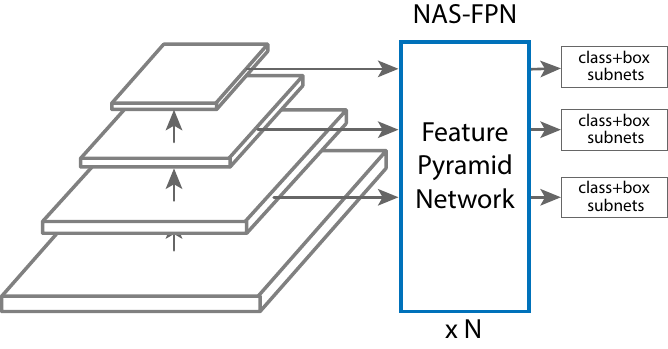}
\caption{RetinaNet with NAS-FPN. In our proposal, feature pyramid network is to be searched by a neural architecture search algorithm. The backbone model and the subnets for class and box predictions follow the original design in RetinaNet~\cite{lin2018focal}. The architecture of FPN can be stacked $N$ times for better accuracy.}
\label{fig:pyramid_network}
\end{figure}

Our method is based on the RetinaNet framework~\cite{lin2018focal} because it is simple and efficient. The RetinaNet framework has two main components: a backbone network (often state-of-the-art image classification network) and a feature pyramid network (FPN).  The goal of the proposed algorithm is to discover a better FPN architecture for RetinaNet. Figure~\ref{fig:pyramid_network} shows the RetinaNet architecture.

To discover a better FPN, we make use of the Neural Architecture Search framework proposed by \cite{zoph2017nas}. The Neural Architecture Search trains a controller to select best model architectures in a given search space using reinforcement learning. The controller uses the accuracy of a child model in the search space as the reward signal to update its parameters. Thus through trial and error the controller learns to generate better architectures over time. As it has been identified by previous works~\cite{tan2018mnasnet,zoph2017nas,zoph2017learning}, the search space plays a crucial role in the success of architecture search. 

In the next section, we design a search space for FPN to generate feature pyramid representations. For scalability of the FPN (i.e., so that an FPN architecture can be stacked repeatedly within RetinaNet), during the search, we also force the the FPN to repeat itself $N$ times and then concatenated into a large architecture. We call our feature pyramid architecture NAS-FPN.

\subsection{Architecture Search Space}
In our search space, a feature pyramid network consists a number of ``merging cells'' that combine a number of input layers into representations for RetinaNet. In the following, we will describe the inputs into the Feature Pyramid Network, and how each merging cell is constructed.
\paragraph{Feature Pyramid Network.}
A feature pyramid network takes multiscale feature layers as inputs and generate output feature layers in the identical scales as shown in Figure \ref{fig:pyramid_network}. We follow the design by RetinaNet \cite{lin2018focal} which uses the last layer in each group of feature layers as the inputs to the first pyramid network. The output of the first pyramid network are the input to the next pyramid network. We use as inputs features in 5 scales $\{C_3, C_4, C_5, C_6, C_7\}$ with corresponding feature stride of $\{8, 16, 32, 64, 128\}$ pixels. The $C_6$ and $C_7$ are created by simply applying stride 2 and stride 4 max pooling to $C_5$. The input features are then passed to a pyramid network consisting of a series of \textit{merging cells} (see below) that introduce cross-scale connections. The pyramid network then outputs augmented multiscale feature representations $\{P_3, P_4, P_5, P_6, P_7\}$. Since both inputs and outputs of a pyramid network are feature layers in the identical scales, the architecture of the FPN can be \emph{stacked} repeatedly for better accuracy. In Section \ref{sec:Experiments}, we show controlling the number of pyramid networks is one simple way to tradeoff detection speed and accuracy.

\paragraph{Merging cell.}
An important observation in previous works in object detection is that it is necessary to ``merge'' features at different scales. The cross-scale connections allow model to combine high-level features with strong semantics and low-level features with high resolution. 

\begin{figure}[h!]
\centering
\includegraphics[width=0.49\textwidth]{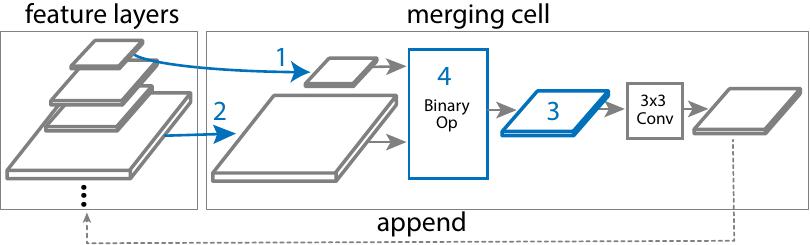}
\caption{Four prediction steps required in a merging cell. Note the output feature layer is pushed back into the stack of candidate feature layers and available for selection for the next merging cell.}
\label{fig:merging_cell}
\end{figure}

We propose \textit{merging cell}, which is a fundamental building block of a FPN, to merge any two input feature layers into a output feature layer. In our implementation, each merging cell takes two input feature layers (could be from different scales), applies processing operations and then combines them to produce one output feature layer of a desired scale. A FPN consists of N different merging cells, where N is given during search. In a merging cell, all feature layers have the same number of filters. The process of constructing a merging cell is shown in Figure~\ref{fig:merging_cell}.

The decisions of how to construct the merging cell are made by a controller RNN. The RNN controller  selects any two candidate feature layers and a binary operation to combine them into a new feature layer, where all  feature layers may have different resolution. Each merging cell has 4 prediction steps made by distinct softmax classifiers:

\vspace{10px}\noindent
\textbf{Step 1.} Select a feature layer $h_i$ from candidates.

\vspace{4px}\noindent
\textbf{Step 2.} Select another feature layer $h_j$ from candidates without replacement.

\vspace{4px}\noindent
\textbf{Step 3.} Select the output feature resolution.

\vspace{4px}\noindent
\textbf{Step 4.} Select a binary op to combine $h_i$ and $h_j$ selected in Step 1 and Step 2 and generate a feature layer with the resolution selected in Step 3.

\begin{figure}[h]
    \centering
    \includegraphics[width=0.25\textwidth]{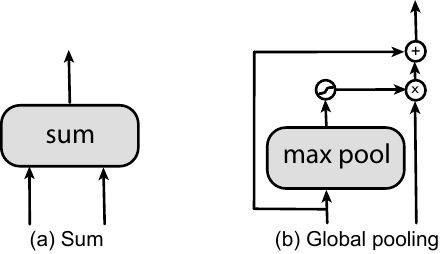}
    \caption{Binary operations.}
    \label{fig:binary_op}
\end{figure}

In step 4, we design two binary operations, \textit{sum} and \textit{global pooling}, in our search space as shown in Figure \ref{fig:binary_op}. These two operations are chosen for their simplicity and efficiency. They do not add any extra trainable parameters. The sum operation is commonly used for combining features~\cite{lin2017fpn}. The design of global pooling operation is inspired by~\cite{wang2018pyramidattention}. We follow Pyramid Attention Networks~\cite{wang2018pyramidattention} except removing convolution layers in the original design. The input feature layers are adjusted to the output resolution by nearest neighbor upsampling or max pooling if needed before applying the binary operation. The merged feature layer is always followed by a ReLU, a 3x3 convolution, and a batch normalization layer.

The input feature layers to a pyramid network form the initial list of input candidates of a merging cell. In Step 5, the newly-generated feature layer is appended to the list of existing input candidates and becomes a new candidate for the next merging cell. There can be multiple candidate features share the same resolution during architecture search. To reduce computation in discovered architecture, we avoid selecting stride 8 feature in Step 3 for intermediate merging cells. In the end, the last 5 merging cells are designed to outputs feature pyramid $\{P_3, P_4, P_5, P_6, P_7\}$. The order of output feature levels is predicted by the controller. Each output feature layer is then generated by repeating the step 1, 2, 4 until the output feature pyramid is fully generated. Similar to~\cite{zoph2017nas}, we take all feature layers that have not
been connected to any of output layer and sum them to the output layer that has the corresponding resolution.

\subsection{Deeply supervised Anytime Object Detection}
One advantage of scaling NAS-FPN with stacked pyramid networks is that the feature pyramid representations can be obtained at output of any given pyramid network. This property enables \textit{anytime detection} which can generate detection results with early exit. Inspired by~\cite{lee2015dsn,huang2018msdn}, we can attach  classifier and box regression heads after all intermediate pyramid networks and train it with deep supervision~\cite{lee2015dsn}. During inference, the model does not need to finish the forward pass for all pyramid networks. Instead, it can stop at the output of any pyramid network and generate detection results. This can be a desirable property when computation resource or latency is a concern and provides a solution that can dynamically decide how much computation resource to allocate for generating detections.
In Appendix~\ref{sec:anytime_detection}, we show NAS-FPN can be used for anytime detection.

\section{Experiments} \label{sec:Experiments}

In this section, we first describe our experiments of Neural Architecture Search to learn a RNN controller to discover the NAS-FPN architecture. Then we demonstrate the discovered NAS-FPN works well with different backbone models and image sizes. The capacity of NAS-FPN can be easily adjusted by changing the number of stacking layers and the feature dimension in pyramid network. We show how to build accurate and fast architectures in the experiments.

\subsection{Implementation Details}

We use the open-source implementation of RetinaNet\footnote{https://github.com/tensorflow/tpu/tree/master/models/official/retinanet}
for experiments. The models are trained on TPUs with 64
images in a batch. During training, we apply multiscale training with a random scale between [0.8, 1.2] to the output image size. The batch normalization layers are applied after all convolution layers.
We use $\alpha=0.25$ and $\gamma=1.5$ for focal loss.
We use a weight decay of 0.0001 and a momentum of 0.9.
The model is trained using 50 epochs. The initial learning rate 0.08 is applied for first 30 epochs and decayed 0.1 at 30 and 40 epochs.
For experiments with DropBlock~\cite{ghiasi2018dropblock},
we use a longer training schedule of 150 epochs with first decay at
120 and the second decay at 140 epochs.
The step-wise learning rate schedule was not stable for training our model with
AmoebaNet backbone on image size of 1280x1280 and for this case we use cosine learning rate schedule. 
The model is trained on COCO train2017 and evaluated
on COCO val2017 for most experiments. In Table~\ref{tab:coco}, we
report test-dev accuracy to compare with existing methods.

\begin{figure}[t]
    \center
    \begin{subfigure}[b]{0.22\textwidth}
        \includegraphics[width=\textwidth]{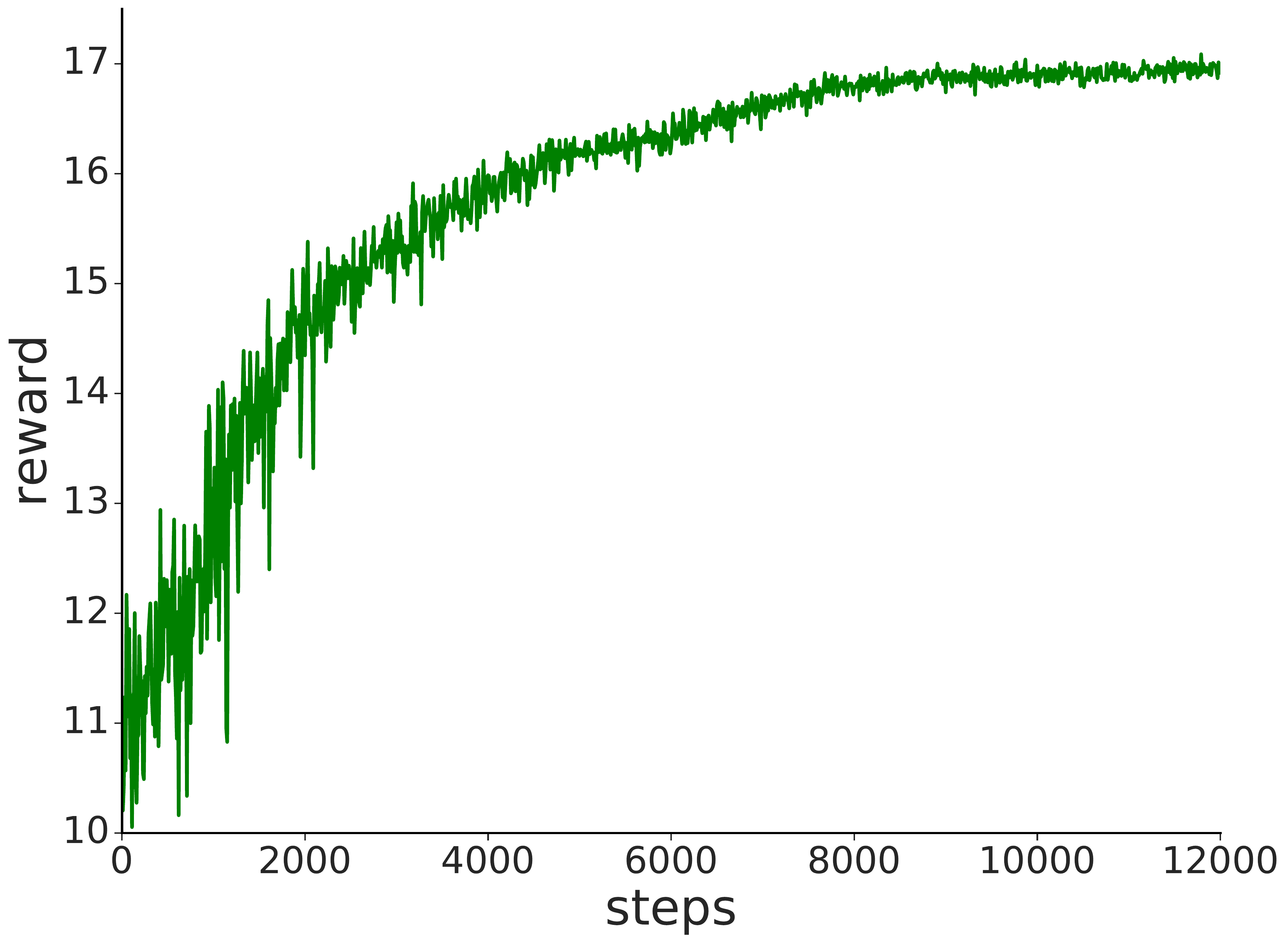} 
    \end{subfigure}
    \begin{subfigure}[b]{0.22\textwidth}
        \includegraphics[width=\textwidth]{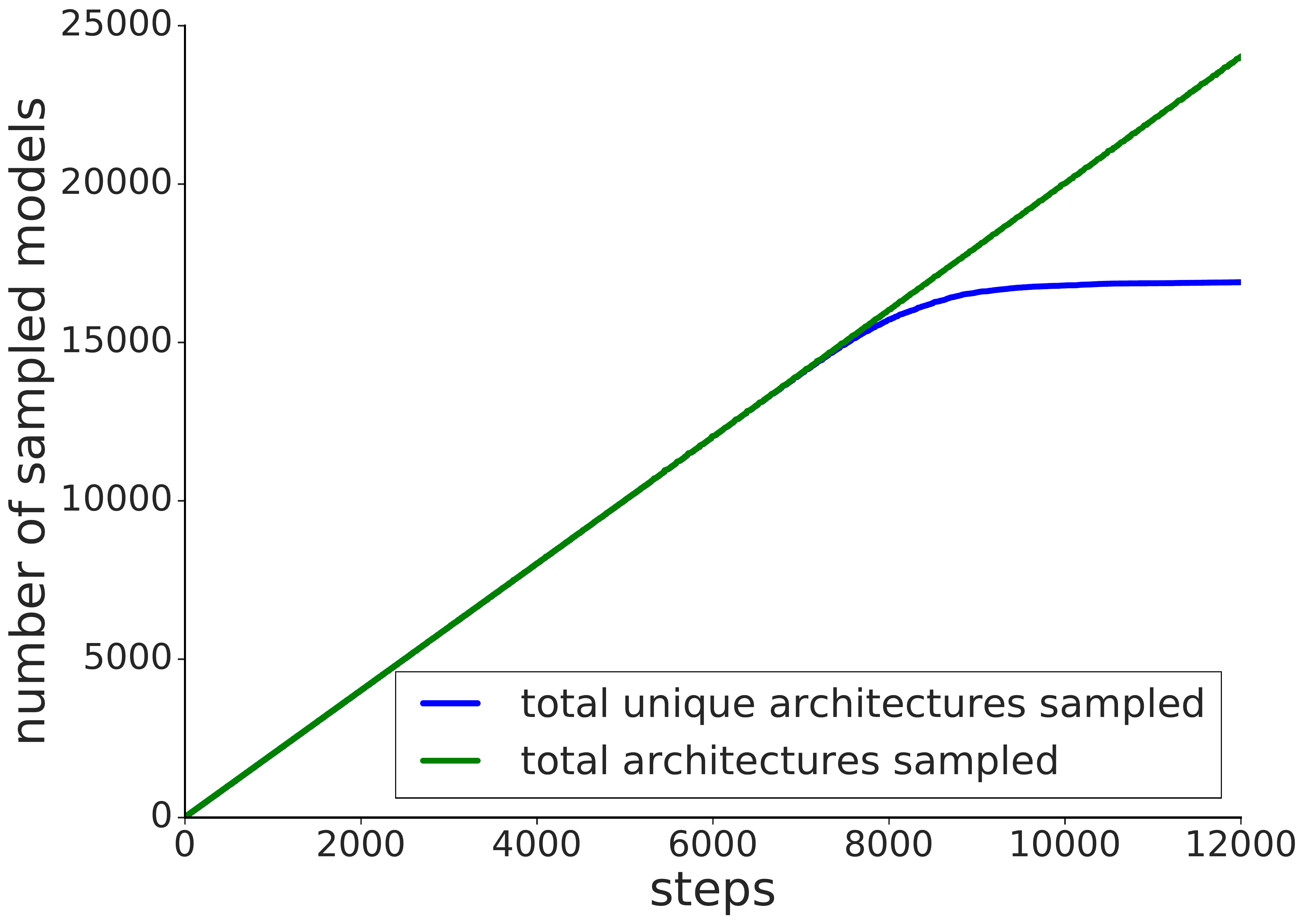} 
    \end{subfigure}
    \caption{Left: Rewards over RL training. The reward is computed as the AP of sampled architectures on the proxy task. Right: The number of sampled unique architectures to the total number of sampled architectures. As controller converges, more identical architectures are sampled by the controller.}
\label{fig:model_search}
\end{figure}

\subsection{Architecture Search for NAS-FPN}

\begin{figure*}[h!]
    \center
    \includegraphics[width=0.9\textwidth]{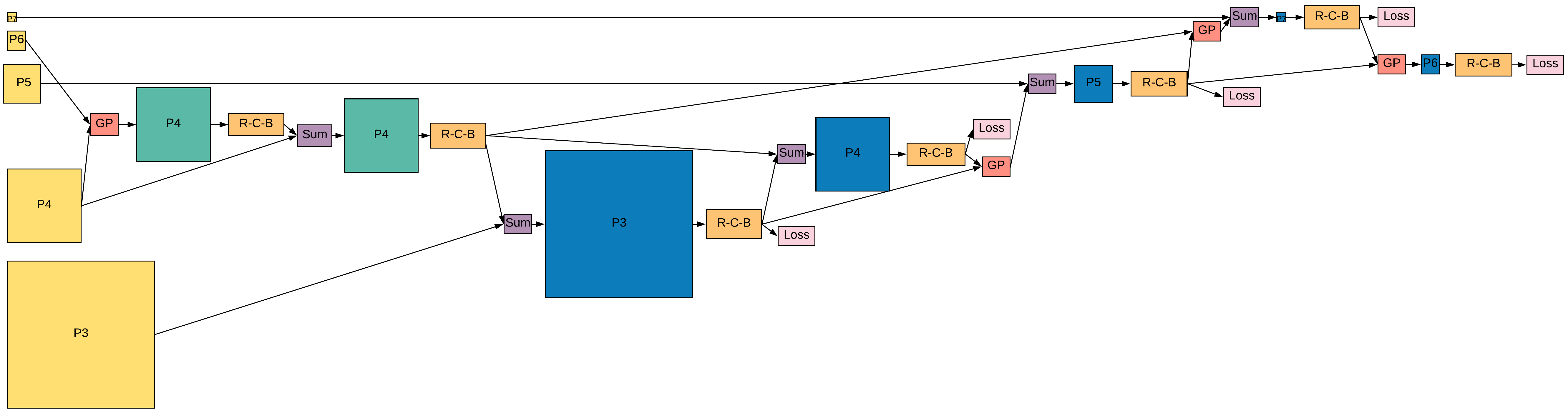}
    \caption{Architecture of the discovered 7-merging-cell pyramid network in NAS-FPN with 5 input layers (yellow) and 5 output feature layers (blue). GP and R-C-B are stands for Global Pooling and ReLU-Conv-BatchNorm, respectively.}
    \label{fig:cell_v1_7nodes}
\end{figure*}
\begin{figure*}[t]
    \center
    \includegraphics[width=0.8\textwidth]{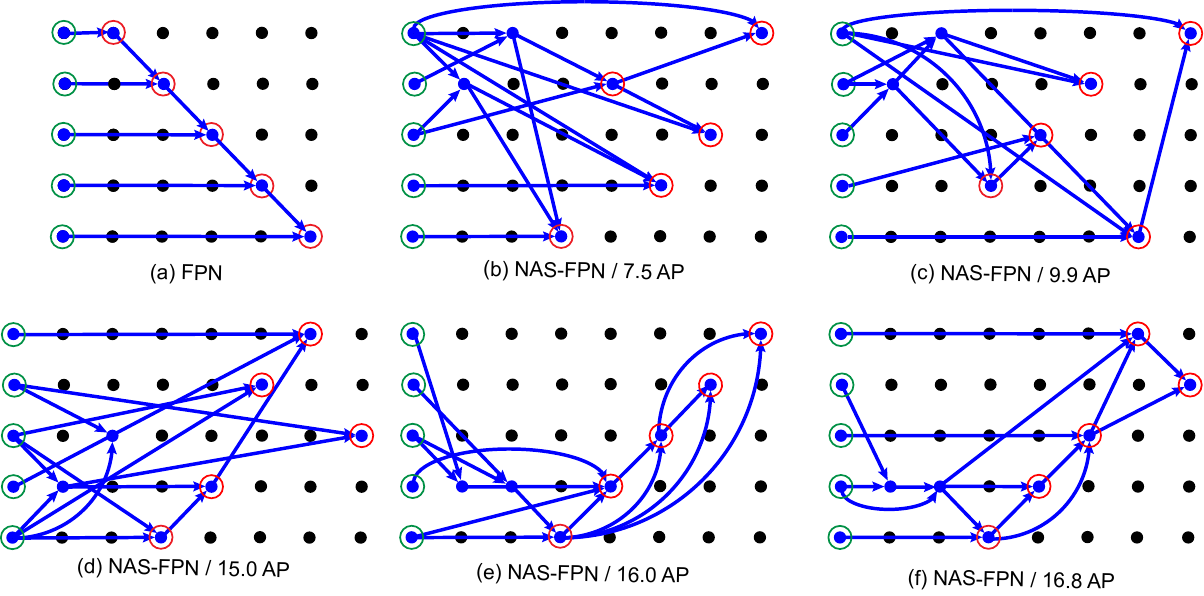} 
    \caption{Architecture graph of NAS-FPN. Each dot represents a feature layer. Feature layers in the same row have identical resolution. The resolution decreases in the bottom-up direction. The arrows indicate the connections between internal layers. The graph is constructed such that an input layer is on the left side. The inputs to a pyramid network are marked with green circles and outputs are marked with red circles.  (a) The baseline FPN architecture. (b-f) The 7-cell NAS-FPN architectures discovered by Neural Architecture Search over training of the RNN controller. The discovered architectures converged as the reward (AP) of the proxy task progressively improves. (f) The final NAS-FPN that we used in our experiments.}
    \label{fig:model_progression}
\end{figure*}

\paragraph{Proxy task.}
To speed up the training of the RNN controller we need a proxy task~\cite{zoph2017learning} that has a short training time and also correlates with the real task. The proxy task can then be used during the search to identify a good FPN architecture. We find that we can simply shorten the training of target task and use it as the proxy task. We only train the proxy task for 10 epochs, instead of 50 epochs that we use to train RetinaNet to converge. To further speed up training proxy task, we use a small backbone architecture of ResNet-10 with input $512\times512$ image size. With these reductions, the training time is ~1hr for a proxy task on TPUs. We repeat the pyramid networks 3 times in our proxy task. The initial learning rate 0.08 is applied for first 8 epochs and decayed by the factor of 0.1 at epoch 8. We reserve a randomly selected 7392 images from the COCO train2017 set as the validation set, which we use to obtain rewards.

\paragraph{Controller.}
Similar  to \cite{zoph2017nas} our controller is a recurrent neural network (RNN) and it is trained using the Proximal Policy Optimization (PPO) \cite{schulman2017proximal} algorithm. The controller samples child networks with different architectures. These architectures are trained on a proxy task using a pool of workers. The workqueue in our experiments consisted of 100 Tensor Processing Units (TPUs). The resulting detection accuracy in average precision (AP) on a held-out validation set is used as the reward to update the controller. Figure \ref{fig:model_search}-Left shows the AP of the sampled architectures for different iterations of training. As it can be seen the controller generated better architectures over time. Figure \ref{fig:model_search}-Right shows total number of sampled architectures and also the total number of unique architectures generated by the RNN controller. The number of unique architectures converged after about 8000 steps. We use the architecture with the highest AP from all sampled architectures during RL training in our experiments. This architecture is first sampled at ~8000 step and sampled many times after that.  Figure \ref{fig:cell_v1_7nodes} shows the details of this architecture.

\begin{figure*}[t]
    \center
    \begin{subfigure}[b]{0.32\textwidth}
        \includegraphics[width=\textwidth]{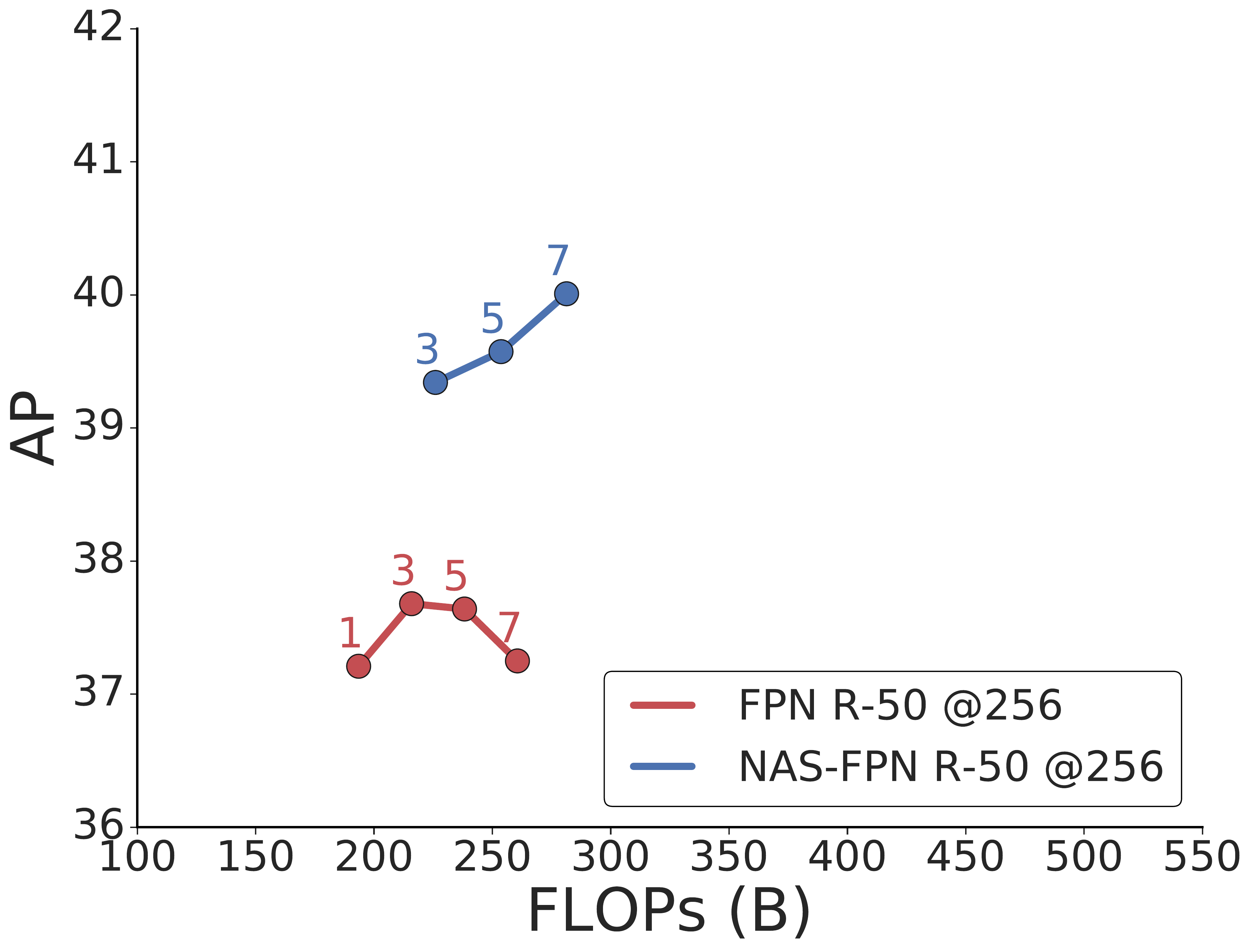} 
        \caption{Number of pyramid networks}
        \label{fig:scale_nasfpn_layers}
    \end{subfigure}
    \begin{subfigure}[b]{0.32\textwidth}
        \includegraphics[width=\textwidth]{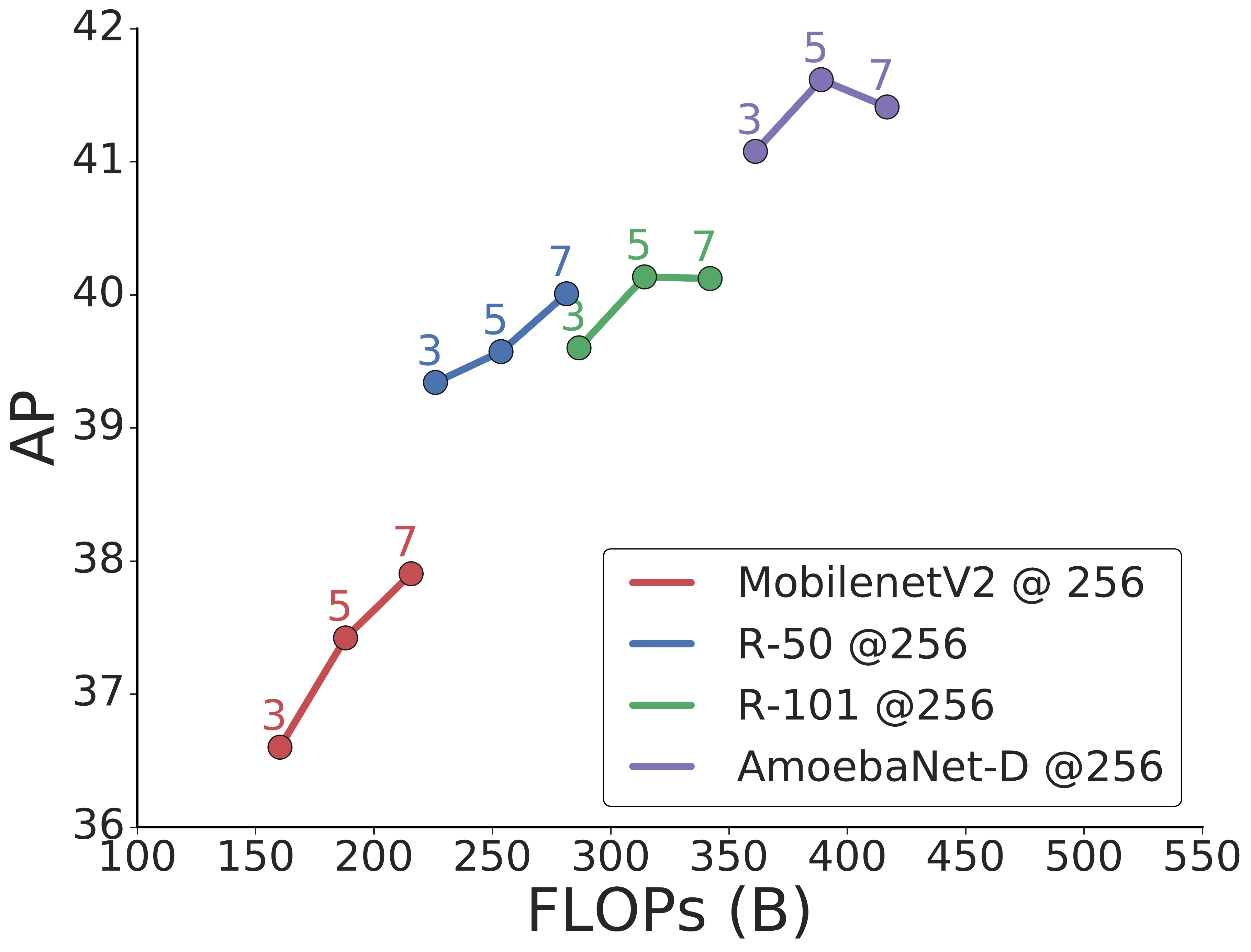} 
        \caption{Backbone architectures}
        \label{fig:scale_nasfpn_backbone}
    \end{subfigure}
    \begin{subfigure}[b]{0.32\textwidth}
        \includegraphics[width=\textwidth]{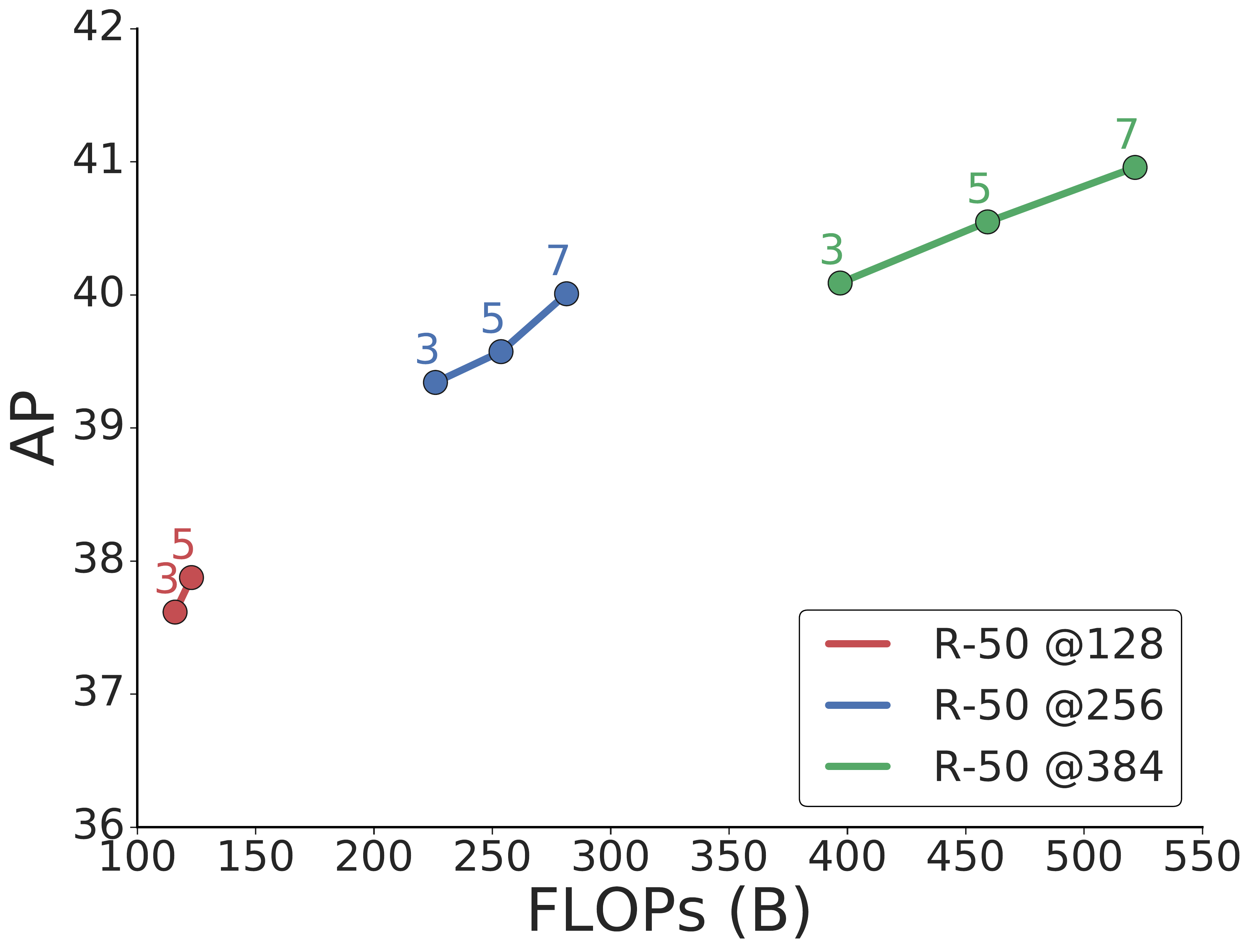} 
        \caption{Feature dimension}
        \label{fig:scale_nasfpn_dim}
    \end{subfigure}
    \caption{The model capacity of NAS-FPN can be controlled with (a) stacking pyramid networks, (b) changing the backbone architecture, and (c) increasing feature dimension in pyramid networks. All models are trained/tested on the image size of 640x640. Number above the marker indicates number of pyramid networks in NAS-FPN.}
\label{fig:scale_nasfpn}
\end{figure*}

\paragraph{Discovered feature pyramid architectures.}
What makes a good feature pyramid architecture? We hope to shed lights on this question by visualizing the discovered architectures. In Figure \ref{fig:model_progression}(b-f), we plot NAS-FPN architectures with progressively higher reward during RL training. We find the RNN controller can quickly pick up some important cross-scale connections in the early learning stage. For example, it discovers the connection between high resolution input and output feature layers, which is critical to generate high resolution features for detecting small objects. As the controller converges, the controller discovers architectures that have both top-down and bottom-up connections which is different from vanilla FPN in Figure \ref{fig:model_progression}(a). We also find better feature reuse as the controller converges. Instead of randomly picking any two input layers from the candidate pool, the controller learns to build connections on newly-generated layers to reuse previously computed feature representations.

\begin{figure*}[t]
    \center
    \begin{subfigure}[b]{1.0\textwidth}
        \centering 
        \includegraphics[width=0.32\linewidth]{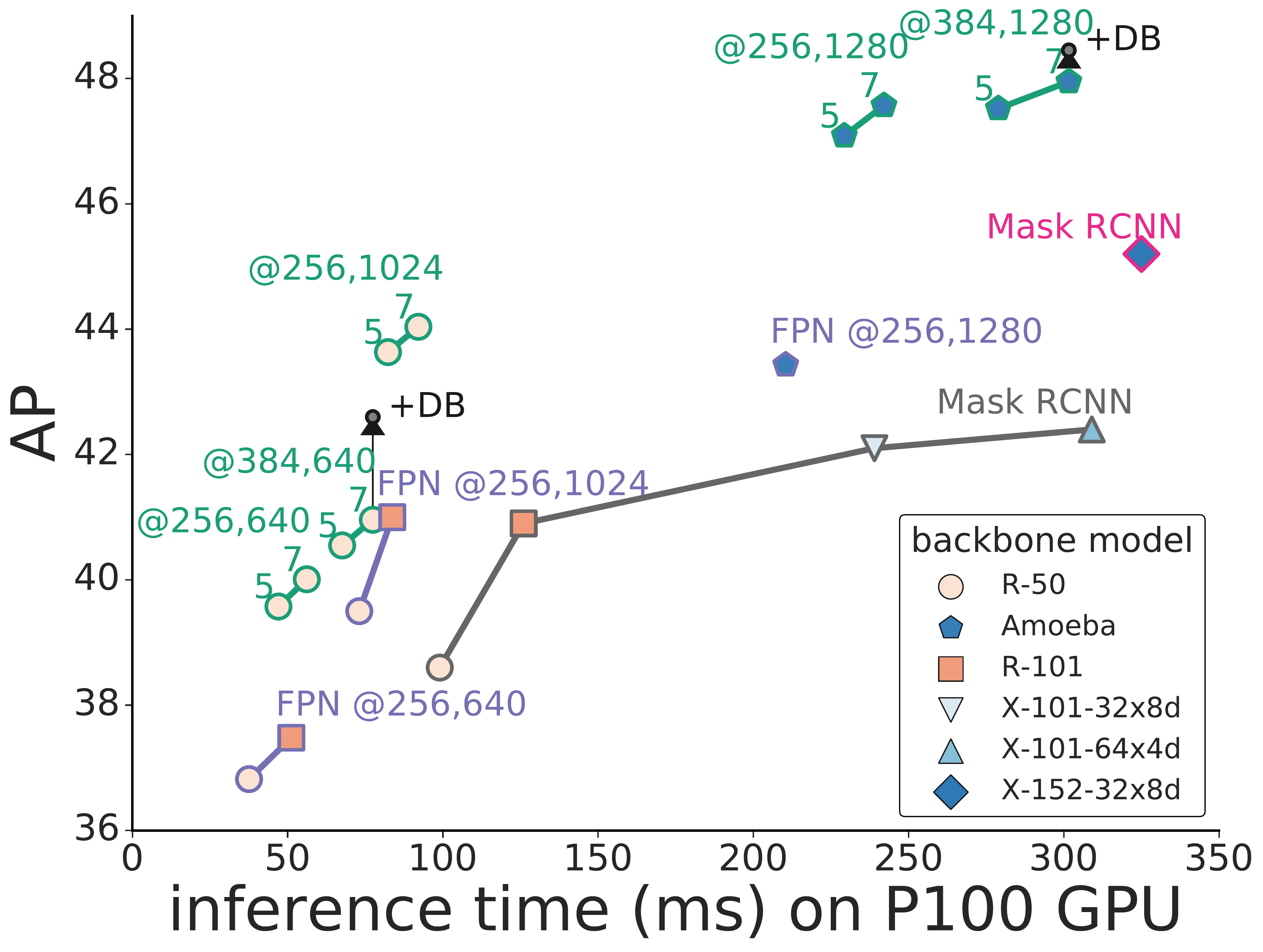} 
        \hfill
        \includegraphics[width=0.32\textwidth]{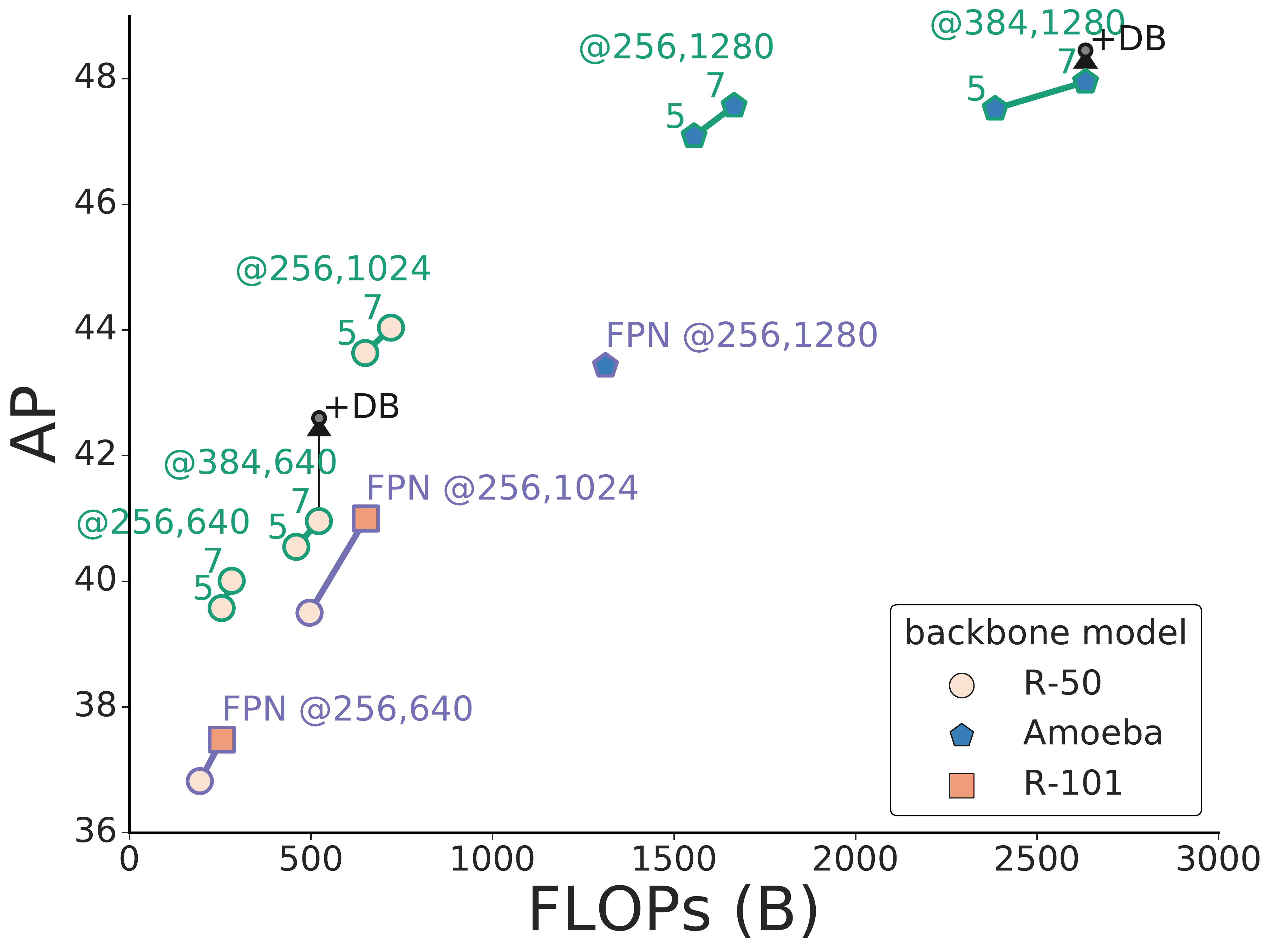} 
        \includegraphics[width=0.32\textwidth]{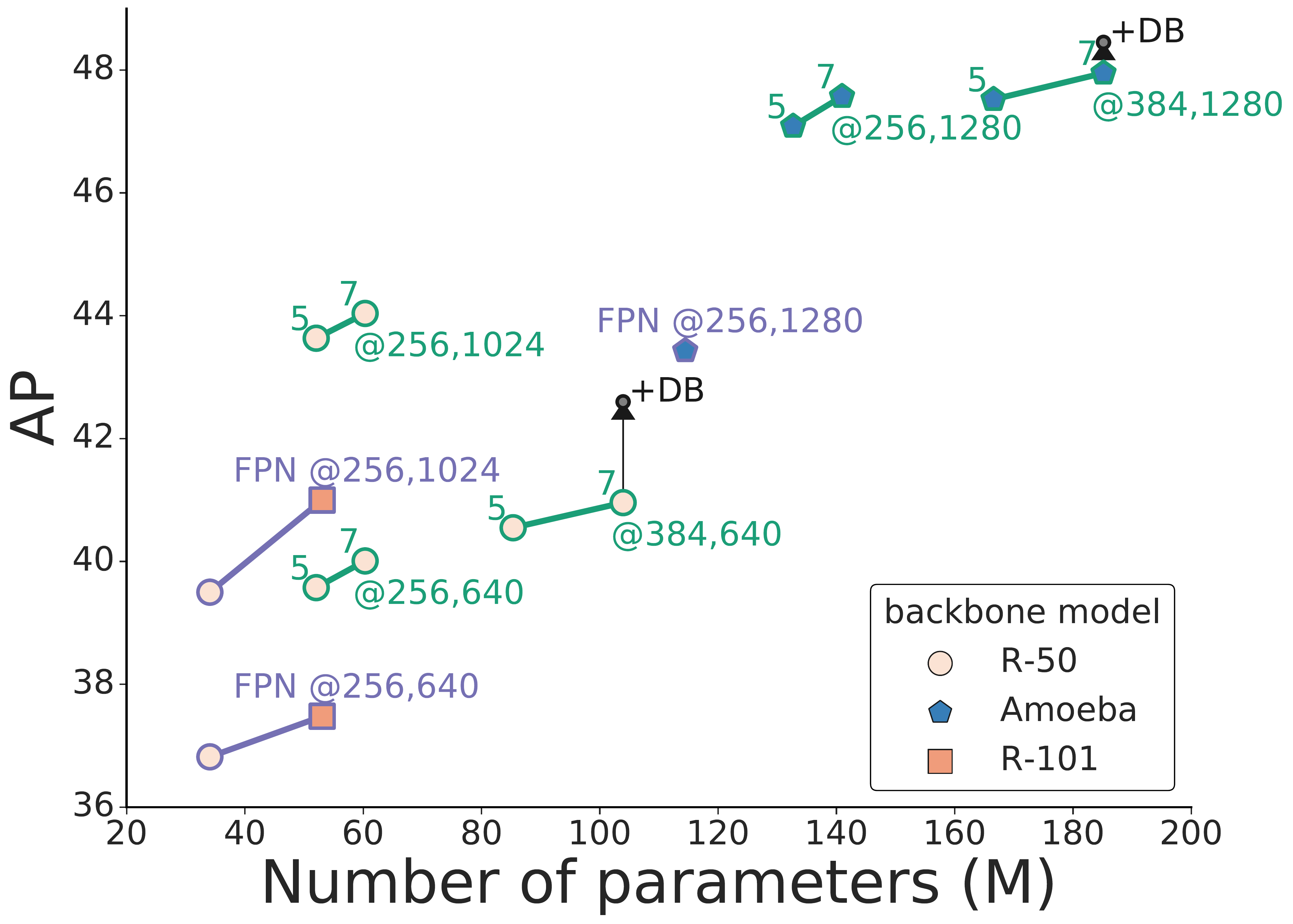} \
        \caption{Accurate models}
        \label{fig:performance_accurate}
    \end{subfigure}
    \begin{subfigure}[b]{1.0\textwidth}
        \centering
        \includegraphics[width=0.32\linewidth]{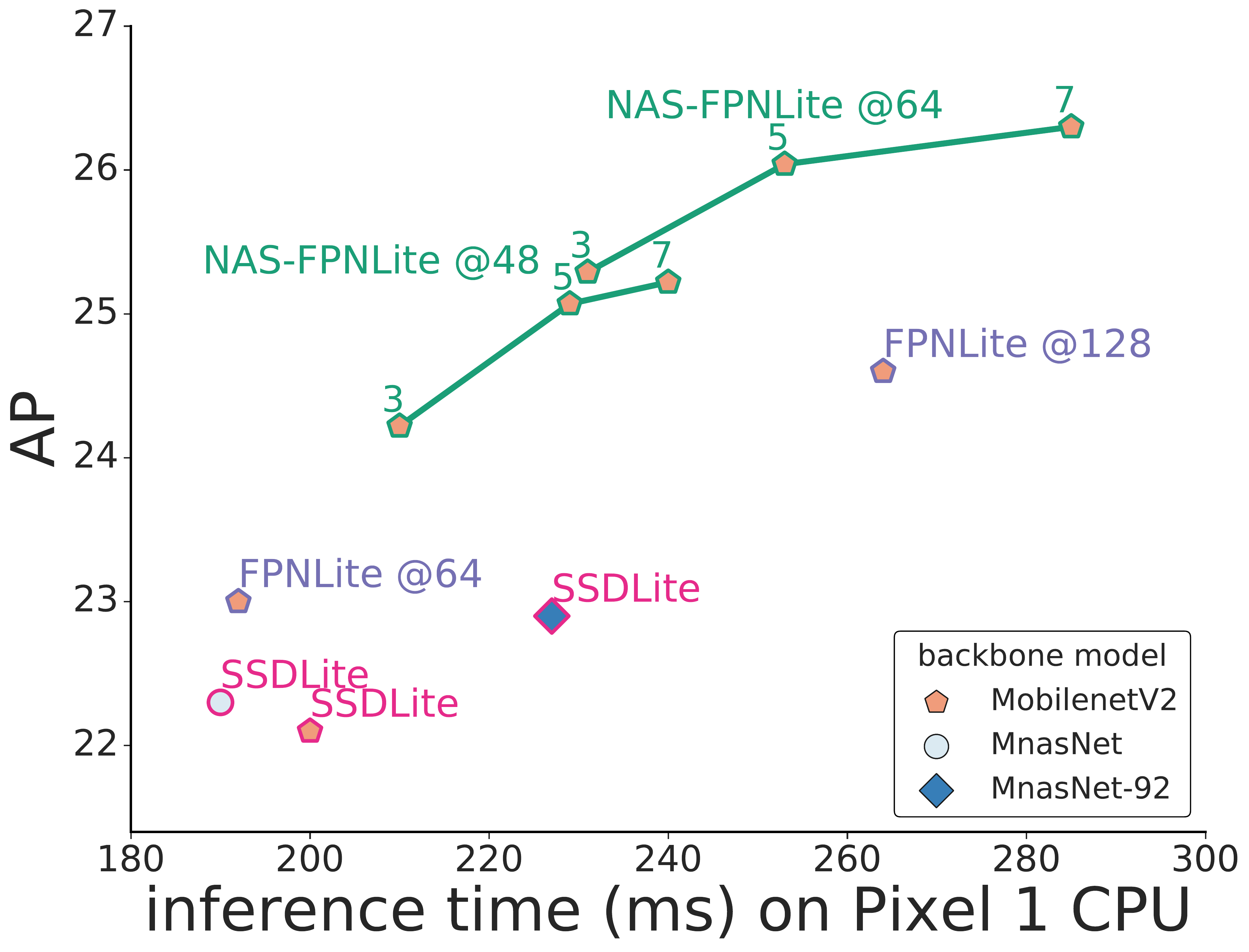} 
        \hfill
        \includegraphics[width=0.32\textwidth]{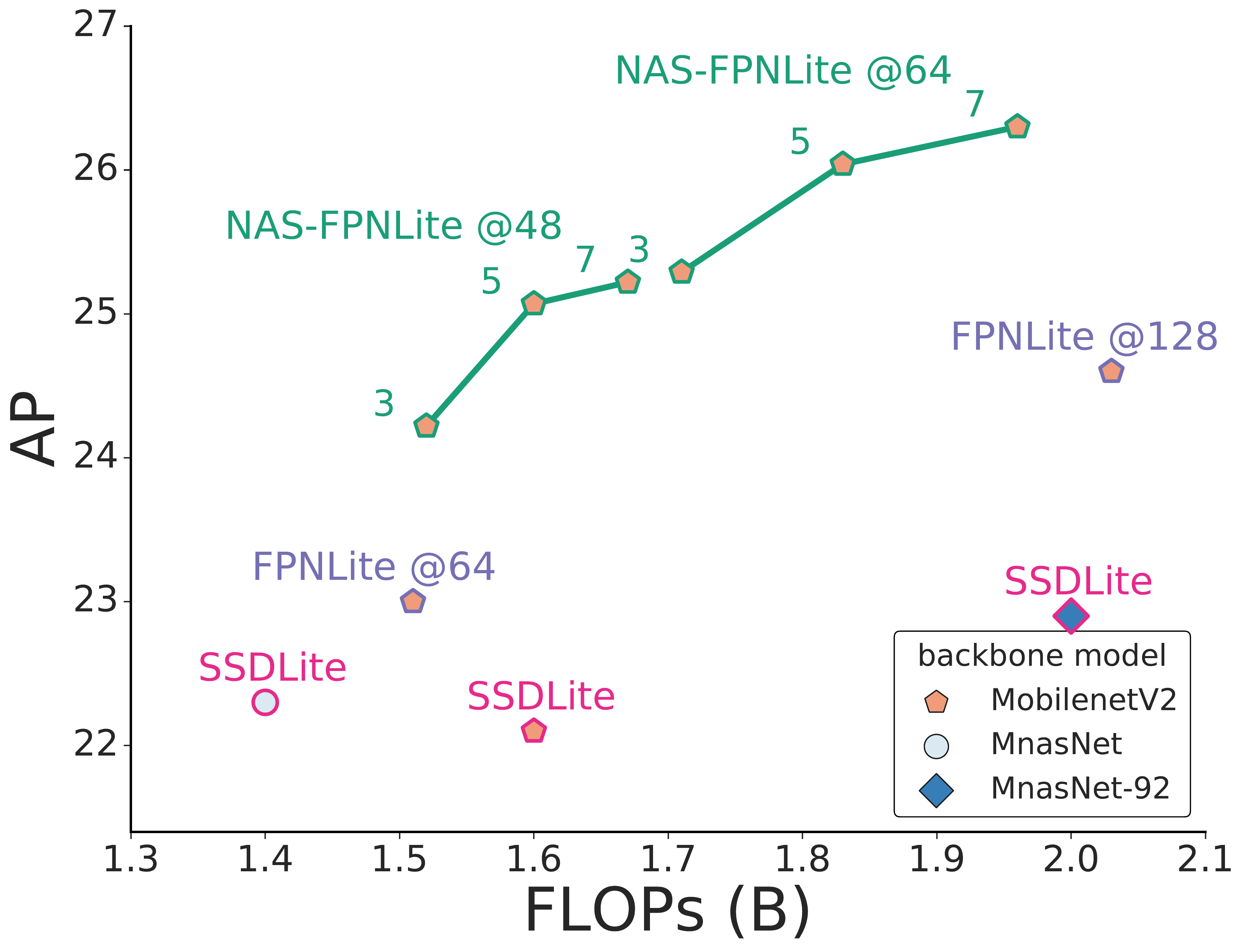} 
        \includegraphics[width=0.32\textwidth]{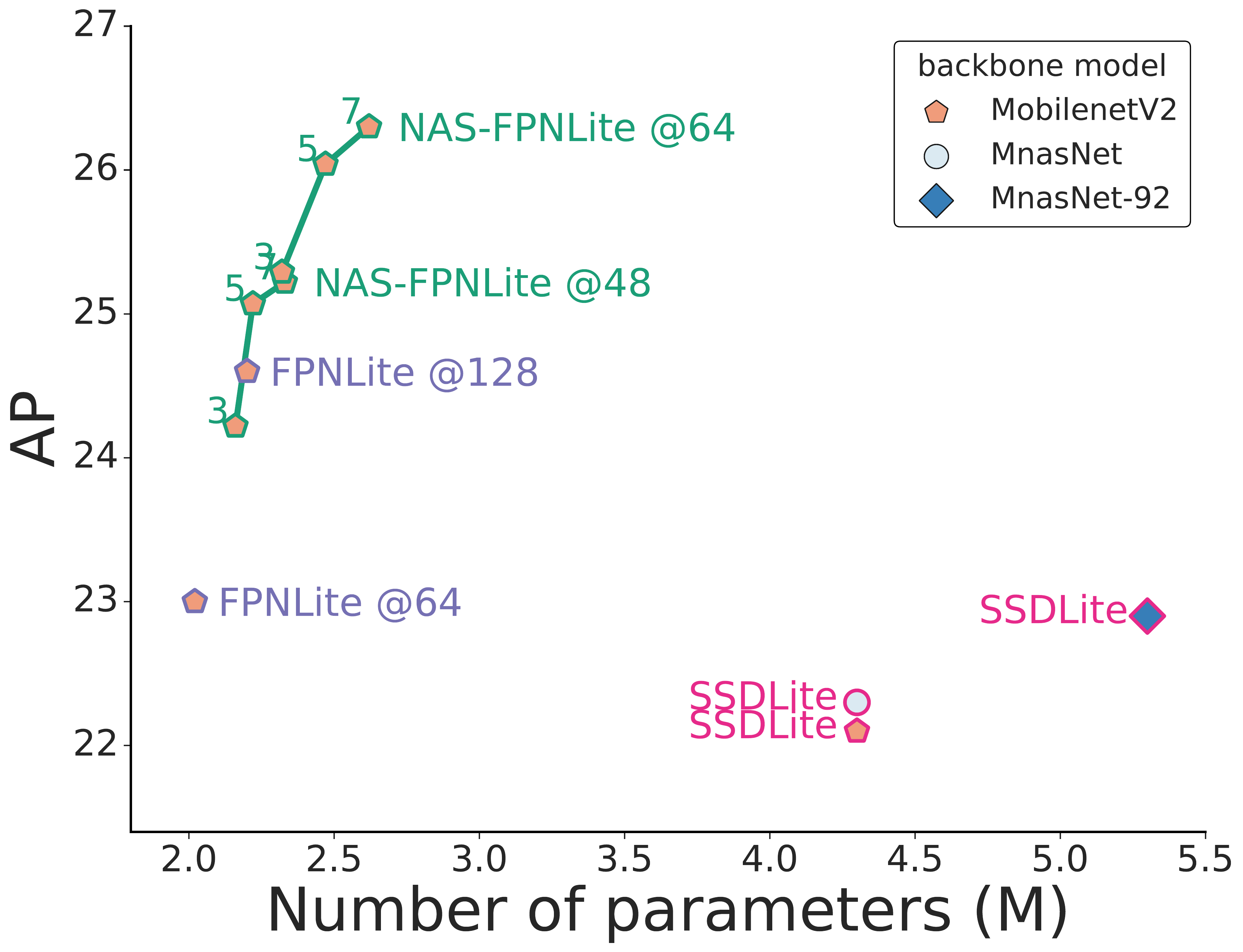} \
        \caption{Fast models}
        \label{fig:performance_fast}
    \end{subfigure}
    \caption{Detection accuracy to inference time (left), FLOPs (middle), and parameters (right). (a) We compare to other high accuracy models. The inference time of all models are computed on a machine with P100 GPU. The green curves highlights results for NAS-FPN with different backbone architectures. The number above the marker indicates the number of repeats of pyramid networks in NAS-FPN. The feature dimension of NAS-FPN/FPN and input image size are mentioned next to each data point. (b) We compare to other fast models. The input image size of all models is 320x320 and the inference times are computed on Pixel 1 CPU. Our model are trained with light-weight model of MobileNetV2.}
    \label{fig:performance}
\end{figure*}

\subsection{Scalable Feature Pyramid Architecture}

In this section, we show how to control the model capacity by adjusting (1) backbone model, (2) the number of repeated pyramid networks, and (3) the number of dimension in pyramid network. We discuss how these adjustments tradeoff computational time and speed.  We define a simple notation to indicate backbone model and NAS-FPN capacity. For example, R-50, 5 @ 256 indicate a model using ResNet-50 backbone model, 5 stacked NAS-FPN pyramid networks, and 256 feature dimension.

\paragraph{Stacking pyramid networks.}
Our pyramid network has a nice property that it can be scaled into a larger architecture by stacking multiple repeated architectures. In Figure \ref{fig:scale_nasfpn_layers}, we show that stacking the vanilla FPN architecture does not always improve performance whereas stacking NAS-FPN improves accuracy significantly. This result highlights our search algorithm can find scalable architectures, which may be hard to design manually. Interestingly, although we only apply 3 pyramid networks for the proxy task during the architecture search phase, the performance still improves with up to 7 pyramid networks applied. 

\paragraph{Adopting different backbone architectures.}
One common way to tradeoff accuracy and speed for object detection architectures is altering the backbone architecture. Although the pyramid network in NAS-FPN was discovered by using a light-weight ResNet-10 backbone architecture, we show that it can be transferred well across different backbone architectures. Figure \ref{fig:scale_nasfpn_backbone} shows the performance of NAS-FPN on top of different backbones, from a lighter weight architecture such as MobilenetV2 to a very high capacity architecture such as AmoebaNet-D~\cite{real2018regularized}. When we apply NAS-FPN with MobilenetV2 on the image size of $640\times 640$, we get 36.6 AP with $160 B$ FLOPs. Using state-of-the-art image classification architecture of AmoebaNet-D \cite{real2018regularized} as the backbone increases the FLOPs to $390 B$ but also adds about 5 AP. NAS-FPN with both light and heavy backbone architectures benefits from stacking more pyramid networks.

\begin{table*}[h!]
\centering
\footnotesize
\begin{tabular}{l|c|ccc|cccc}
\toprule
\multicolumn{1}{c|}{\bf model} & \bf image size & \multicolumn{1}{l}{\bf \# FLOPs} & \multicolumn{1}{l}{\bf \# params} & \multicolumn{1}{c|}{\bf inference time (ms)}  &
\multicolumn{1}{l}\bf{test-dev AP}
\\ \midrule
YOLOv3    DarkNet-53 \cite{redmon2018yolov3}&  $320\times320$& 38.97 B & - & 22 (Titan X) &  28.2 \\

MobileNetV2 + SSDLite \cite{tan2018mnasnet}&  $320\times320$& 1.6B & 4.3M & 200 (Pixel 1 CPU) &  22.1\\

MnasNet + SSDLite \cite{tan2018mnasnet}&  $320\times320$& 1.4B & 4.3M & 190 (Pixel 1 CPU) &  22.3 \\
MnasNet-92 + SSDLite \cite{tan2018mnasnet}&  $320\times320$& 2.0B & 5.3M & 227 (Pixel 1 CPU) &  22.9 \\
\midrule
FPNLite MobileNetV2\; @ 64  & $320\times 320$  & 1.51B  & 2.02M           & 192 (Pixel 1 CPU) & 22.7 \\
FPNLite MobileNetV2\; @ 128  & $320\times 320$  & 2.03B  & 2.20M           & 264 (Pixel 1 CPU) & 24.3 \\

\midrule

NAS-FPNLite MobileNetV2\;(3 @ 48)  & $320\times 320$  & 1.52 B  & 2.16 M           & 210 (Pixel 1 CPU) & 24.2 \\
NAS-FPNLite MobileNetV2\;(7 @ 64)  & $320\times 320$  & 1.96 B  & 2.62 M           & 285 (Pixel 1 CPU) & 25.7 \\
\midrule
\midrule
\midrule
YOLOv3    DarkNet-53 \cite{redmon2018yolov3}&  $608\times608$& 140.69 B & - & 51 (Titan X) &  33.0 \\
CornerNet Hourglass \cite{law2018cornernet}& $512\times 512$ & - & - & 244 (Titan X) & 40.5 \\
Mask R-CNN X-152-32x8d \cite{he2017mask} & $1280\times800$ & - &  - & 325 (P100) & 45.2  
\\
RefineDet R-101 \cite{zhang2018refinedet}& $832\times 500$     & - & - & 90 (Titan X) & 34.4 \\
\midrule
FPN R-50 @256 \cite{lin2018focal} & $640 \times 640$ & 193.6B & 34.0M  & 37.5 (P100) & 37.0 \\
FPN R-101 @256 \cite{lin2018focal} & $640 \times 640$ & 254.2B & 53.0M  & 51.1 (P100)  & 37.8 \\
FPN R-50 @256 \cite{lin2018focal} & $1024 \times 1024$ & 495.8B & 34.0M  & 73.0 (P100) & 40.1 \\
FPN R-101 @256 \cite{lin2018focal} & $1024 \times 1024$ & 651.1B & 53.0M  & 83.7 (P100)  & 41.1 \\
FPN AmoebaNet @256 \cite{lin2018focal} & $1280 \times 1280$ & 1311 B & 114.4 M  & 210.4 (P100)  & 43.4 \\

\midrule

NAS-FPN R-50\;(7 @ 256) & $640 \times 640$ & 281.3B             & 60.3M  & 56.1 (P100)  & 39.9 \\
NAS-FPN R-50\;(7 @ 256) & $1024 \times 1024$ & 720.4B & 60.3M & 92.1 (P100)  & 44.2 \\
NAS-FPN R-50\;(7 @ 256) & $1280 \times 1280$ & 1125.5B & 60.3M & 131.9 (P100)  & 44.8 \\
NAS-FPN R-50\;(7 @ 384) & $1280 \times 1280$ & 2086.3B & 103.9 M & 192.3 (P100)  & 45.4 \\
NAS-FPN R-50\;(7 @ 384) + DropBlock & $1280 \times 1280$ & 2086.3B & 103.9M & 192.3 (P100)  & 46.6 \\
\midrule
NAS-FPN AmoebaNet\;(7 @ 384) & $1280\times 1280$ & 2633 B & 166.5 M & 278.9 (P100) & 48.0 \\ 
NAS-FPN AmoebaNet\;(7 @ 384) + DropBlock & $1280 \times 1280$ & 2633 B & 166.5 M & 278.9 (P100) & 48.3 \\
\bottomrule
\end{tabular}
\vspace{0.2cm}
\caption{Performance of RetinaNet with NAS-FPN and other state-of-the-art detectors on test-dev set of COCO.}
\label{tab:coco}
\end{table*}

\paragraph{Adjusting feature dimension of feature pyramid networks.}
Another way to increase the capacity of a model is to increase the feature dimension of feature layers in NAS-FPN. Figure \ref{fig:scale_nasfpn_dim} shows results of 128, 256, and 384 feature dimension in NAS-FPN with a ResNet-50 backbone architecture. Not surprisingly, increasing the feature dimension improves detection performance but it may not be an efficient way to improve the performance. In Figure \ref{fig:scale_nasfpn_dim}, R-50 7 @ 256, with much less FLOPs, achieves similar AP compared to R-50 3 @ 384. Increasing feature dimension would require model regularization technique. In Section \ref{sec:regularization}, we discuss using DropBlock~\cite{ghiasi2018dropblock} to regularize the model.

\paragraph{Architectures for high detection accuracy.}
With the scalable NAS-FPN architecture, we discuss how to build an accurate model while remaining efficient. In Figure \ref{fig:performance_accurate}, we first show that NAS-FPN R-50 5 @256 model has comparable FLOPs to the R-101 FPN baseline but with 2.5 AP gain. This shows using NA S-FPN is more effective than replacing the backbone with a higher capacity model. Going for a higher accuracy model, one can use a heavier backbone model or higher feature dimensions. Figure~\ref{fig:performance_accurate} shows that NAS-FPN architectures are in the upper left part in the accuracy to inference time figure compared to existing methods. The NAS-FPN is as accurate as to the state-of-the-art Mask R-CNN model with less computation time.

\paragraph{Architectures for fast inference.}
Designing object detector with low latency and limited computation budget is an active research topic. Here, we introduce NAS-FPNLite for mobile object detection. The major difference of NAS-FPNLite and NAS-FPN is that we search a pyramid network that has outputs from $P_3$ to $P_6$. Also we follow SSDLite~\cite{sandler2018mobilenetv2} and replace convolution with depth-wise separable convolution in NAS-FPN. We discover  a 15-cell architecture which yields good performance and use it in our experiments. We combine NAS-FPNLite and  MobileNetV2~\cite{sandler2018mobilenetv2} in RetinaNet framework. For a fair comparison, we create a FPNLite baseline, which follows the original FPN structure and replaces all convolution layers with depth-wise separable convolution.
Following \cite{tan2018mnasnet, sandler2018mobilenetv2}, we train NAS-FPNLite and FPNLite using an open-source object detection API.\footnote{https://github.com/tensorflow/models/tree/master/research/object\_detection}
In Figure \ref{fig:performance_fast}, we control the feature dimension of NAS-FPN to be 48 or 64 so that it has similar FLOPs and CPU runtime on Pixel 1 as baseline methods and show that NAS-FPNLite outperforms both SSDLite~\cite{sandler2018mobilenetv2} and FPNLite.

\subsection{Further Improvements with DropBlock} \label{sec:regularization}
\begin{figure}[h!]
\center
\includegraphics[width=0.35\textwidth]{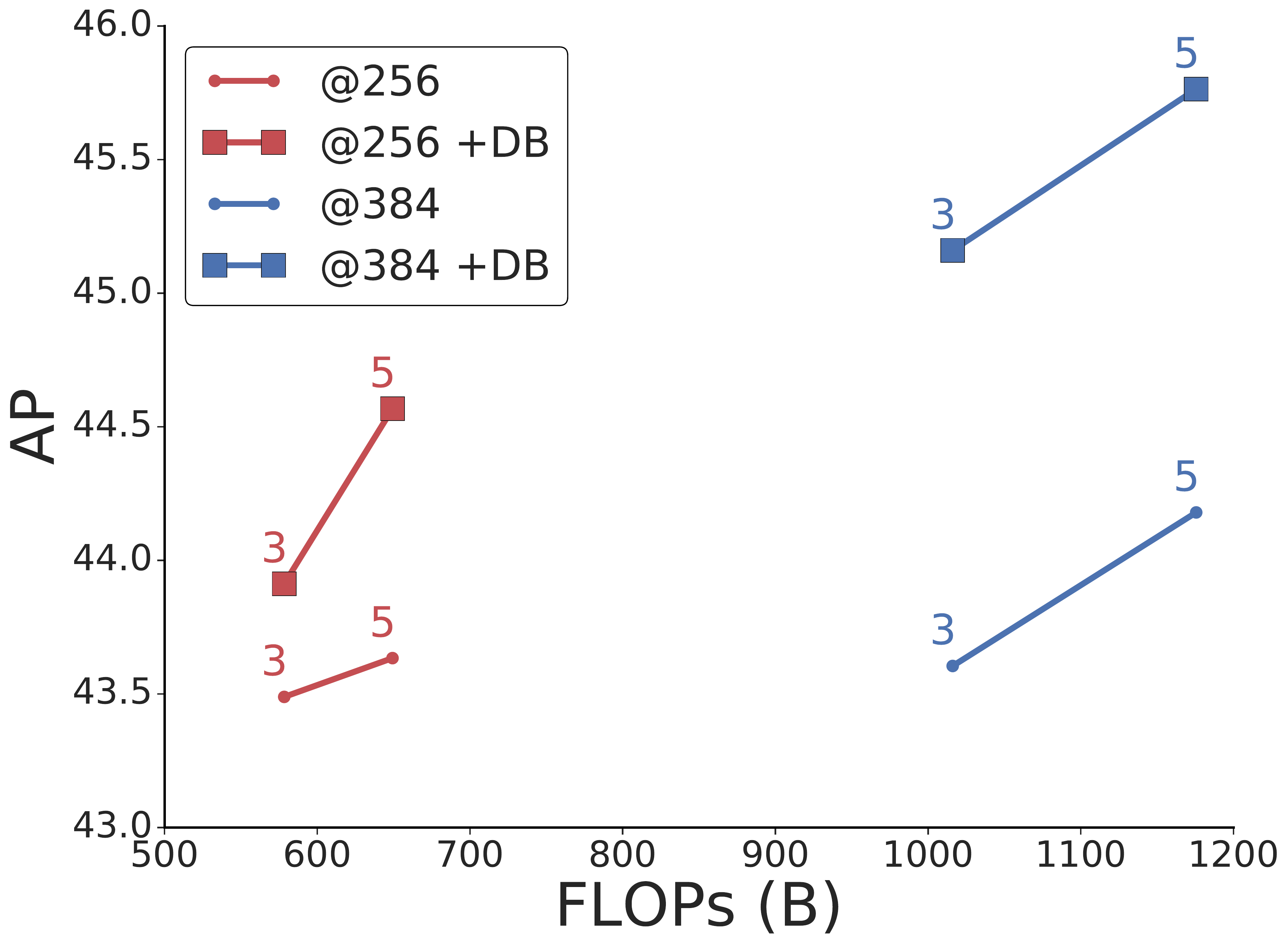}
\caption{Performance comparison of NAS-FPN with feature dimension of 256
or 384 when it is trained with and without DropBlock (DB). Models are trained with backbone of ResNet-50 on image size of 1024x1024.
Adding DropBlock is more important when we increase feature dimension in pyramid networks.}
\label{fig:dropblock}
\end{figure}

Due to the increased number of new layers introduced in NAS-FPN architecture, a proper model regularization is needed to prevent overfitting. Following the technique in \cite{ghiasi2018dropblock}, we apply DropBlock with block size 3x3 after batch normalization layers in the the NAS-FPN layers. Figure \ref{fig:dropblock} shows DropBlock improves the performance of NAS-FPN. Especially, it improves more for architecture that has more newly introduced filters. Note that by default we do not apply DropBlock in previous experiments for the fair comparison to existing works.

\section{Conclusion}
In this paper, we proposed to use Neural Architecture Search to further optimize the process of designing Feature Pyramid Networks for Object Detection. Our experiments on the COCO dataset showed that the discovered architecture, named NAS-FPN, is flexible and performant for building accurate detection model. On a wide range of accuracy and speed tradeoff, NAS-FPN produces significant improvements upon many backbone architectures.

{\small
\bibliographystyle{ieee}
\bibliography{egbib}

\begin{thebibliography}{10}\itemsep=-1pt

\bibitem{adelson1984pyramid}
E.~H. Adelson, C.~H. Anderson, J.~R. Bergen, P.~J. Burt, and J.~M. Ogden.
\newblock Pyramid methods in image processinh.
\newblock {\em RCA engineer}, 1984.

\bibitem{baker2017designing}
B.~Baker, O.~Gupta, N.~Naik, and R.~Raskar.
\newblock Designing neural network architectures using reinforcement learning.
\newblock In {\em ICLR}, 2016.

\bibitem{bolukbasi2017adaptive}
T.~Bolukbasi, J.~Wang, O.~Dekel, and V.~Saligrama.
\newblock Adaptive neural networks for efficient inference.
\newblock In {\em ICML}, 2017.

\bibitem{chen2018searching}
L.-C. Chen, M.~D. Collins, Y.~Zhu, G.~Papandreou, B.~Zoph, F.~Schroff, H.~Adam,
  and J.~Shlens.
\newblock Searching for efficient multi-scale architectures for dense image
  prediction.
\newblock In {\em NIPS}, 2018.

\bibitem{rubio2018shortcut}
R.~J. L.-S. D.~Oñoro-Rubio, M.~Niepert.
\newblock Learning short-cut connections for object counting.
\newblock {\em BMVC}, 2018.

\bibitem{elsken2018neural}
T.~Elsken, J.~H. Metzen, and F.~Hutter.
\newblock Neural architecture search: A survey.
\newblock {\em arXiv preprint arXiv:1808.05377}, 2018.

\bibitem{fu2017dssd}
C.~Fu, W.~Liu, A.~Ranga, A.~Tyagi, and A.~C. Berg.
\newblock {DSSD} : Deconvolutional single shot detector.
\newblock {\em CoRR}, abs/1701.06659, 2017.

\bibitem{ghiasi2016pyramid}
G.~Ghiasi and C.~C. Fowlkes.
\newblock Laplacian pyramid reconstruction and refinement for semantic
  segmentation.
\newblock In {\em ECCV}, 2016.

\bibitem{ghiasi2018dropblock}
G.~Ghiasi, T.~Lin, and Q.~V. Le.
\newblock {DropBlock}: A regularization method for convolutional networks.
\newblock {\em NIPS}, 2018.

\bibitem{Detectron2018}
R.~Girshick, I.~Radosavovic, G.~Gkioxari, P.~Doll\'{a}r, and K.~He.
\newblock Detectron.
\newblock \url{https://github.com/facebookresearch/detectron}, 2018.

\bibitem{he2017mask}
K.~He, G.~Gkioxari, P.~Doll{\'a}r, and R.~Girshick.
\newblock {Mask R-CNN}.
\newblock In {\em ICCV}, 2017.

\bibitem{he2016resnet}
K.~He, X.~Zhang, S.~Ren, and J.~Sun.
\newblock Deep residual learning for image recognition.
\newblock In {\em CVPR}, 2016.

\bibitem{huang2018msdn}
G.~Huang, D.~Chen, T.~Li, F.~Wu, L.~{van der Maaten}, and K.~Weinberger.
\newblock Multi-scale dense networks for resource efficient image
  classification.
\newblock In {\em ICLR}, 2018.

\bibitem{huang2017multi}
G.~Huang, D.~Chen, T.~Li, F.~Wu, L.~van~der Maaten, and K.~Q. Weinberger.
\newblock Multi-scale dense networks for resource efficient image
  classification.
\newblock In {\em ICLR}, 2017.

\bibitem{huang2017densenet}
G.~Huang, Z.~Liu, and K.~Q. Weinberger.
\newblock Densely connected convolutional networks.
\newblock In {\em CVPR}, 2017.

\bibitem{liu2018reconfigfpn}
T.~Kong, F.~Sun, W.~Huang, and H.~Liu.
\newblock Deep feature pyramid reconfiguration for object detection.
\newblock In {\em ECCV}, 2018.

\bibitem{kong2016ron}
T.~Kong, F.~Sun, A.~Yao, H.~Liu, M.~Lu, and Y.~Chen.
\newblock {RON:} reverse connection with objectness prior networks for object
  detection.
\newblock In {\em CVPR}, 2017.

\bibitem{law2018cornernet}
H.~Law and J.~Deng.
\newblock Cornernet: Detecting objects as paired keypoints.
\newblock In {\em ECCV}, 2018.

\bibitem{lee2015dsn}
C.-Y. Lee, S.~Xie, P.~Gallagher, Z.~Zhang, and Z.~Tu.
\newblock Deeply-supervised nets.
\newblock In {\em AISTATS}, 2015.

\bibitem{wang2018pyramidattention}
H.~Li, P.~Xiong, J.~An, and L.~Wang.
\newblock Pyramid attention network for semantic segmentation.
\newblock {\em BMVC}, 2018.

\bibitem{li2018detnet}
Z.~Li, C.~Peng, G.~Yu, X.~Zhang, Y.~Deng, and J.~Sun.
\newblock Detnet: A backbone network for object detection.
\newblock In {\em ECCV}, 2018.

\bibitem{lin2017fpn}
T.-Y. Lin, P.~Doll{\'a}r, R.~B. Girshick, K.~He, B.~Hariharan, and S.~J.
  Belongie.
\newblock Feature pyramid networks for object detection.
\newblock In {\em CVPR}, 2017.

\bibitem{lin2018focal}
T.-Y. Lin, P.~Goyal, R.~Girshick, K.~He, and P.~Doll{\'a}r.
\newblock Focal loss for dense object detection.
\newblock In {\em ICCV}, 2017.

\bibitem{liu2017progressive}
C.~Liu, B.~Zoph, J.~Shlens, W.~Hua, L.-J. Li, L.~Fei-Fei, A.~Yuille, J.~Huang,
  and K.~Murphy.
\newblock Progressive neural architecture search.
\newblock In {\em ECCV}, 2017.

\bibitem{liu2018panet}
S.~Liu, L.~Qi, H.~Qin, J.~Shi, and J.~Jia.
\newblock Path aggregation network for instance segmentation.
\newblock In {\em CVPR}, 2018.

\bibitem{liu2016ssd}
W.~Liu, D.~Anguelov, D.~Erhan, C.~Szegedy, S.~Reed, C.-Y. Fu, and A.~C. Berg.
\newblock {SSD:} single shot multibox detector.
\newblock In {\em ECCV}, 2016.

\bibitem{islam2017gatednetwork}
N.~D. B.~B. Md~Amirul~Islam, Mrigank~Rochan and Y.~Wang.
\newblock Gated feedback refinement network for dense image labeling.
\newblock {\em CVPR}, 2017.

\bibitem{newell2016hourglass}
A.~Newell, K.~Yang, and J.~Deng.
\newblock Stacked hourglass networks for human pose estimation.
\newblock In {\em ECCV}, 2016.

\bibitem{real2018regularized}
E.~Real, A.~Aggarwal, Y.~Huang, and Q.~V. Le.
\newblock Regularized evolution for image classifier architecture search.
\newblock In {\em AAAI}, 2018.

\bibitem{redmon2018yolov3}
J.~Redmon and A.~Farhadi.
\newblock Yolov3: An incremental improvement.
\newblock {\em arXiv preprint arXiv:1804.02767}, 2018.

\bibitem{ronneberger2015unet}
O.~Ronneberger, P.~Fischer, and T.~Brox.
\newblock {U-Net}: Convolutional networks for biomedical image segmentation.
\newblock In {\em Medical Image Computing and Computer-Assisted Intervention},
  2015.

\bibitem{sandler2018mobilenetv2}
M.~Sandler, A.~Howard, M.~Zhu, A.~Zhmoginov, and L.-C. Chen.
\newblock {MobileNetV2:} inverted residuals and linear bottl.
\newblock {\em CVPR}, 2019.

\bibitem{schulman2017proximal}
J.~Schulman, F.~Wolski, P.~Dhariwal, A.~Radford, and O.~Klimov.
\newblock Proximal policy optimization algorithms.
\newblock {\em arXiv preprint arXiv:1707.06347}, 2017.

\bibitem{kim2018pfpnet}
J.-Y. S. M.-C. K. S.-J.~K. Seung-Wook~Kim, Hyong-Keun~Kook.
\newblock Parallel feature pyramid network for object detection.
\newblock {\em ECCV}, 2018.

\bibitem{szegedy2015googlenet}
C.~Szegedy, W.~Liu, Y.~Jia, P.~Sermanet, S.~E. Reed, D.~Anguelov, D.~Erhan,
  V.~Vanhoucke, and A.~Rabinovich.
\newblock Deep residual learning for image recognition.
\newblock In {\em CVPR}, 2015.

\bibitem{tan2018mnasnet}
M.~Tan, B.~Chen, R.~Pang, V.~Vasudevan, and Q.~V. Le.
\newblock Mnasnet: Platform-aware neural architecture search for mobile.
\newblock {\em arXiv preprint arXiv:1807.11626}, 2018.

\bibitem{teerapittayanon2016branchynet}
S.~Teerapittayanon, B.~McDanel, and H.~Kung.
\newblock Branchynet: Fast inference via early exiting from deep neural
  networks.
\newblock In {\em ICPR}, pages 2464--2469. IEEE, 2016.

\bibitem{woo2018stairnet}
S.~Woo, S.~Hwang, and I.~S. Kweon.
\newblock {StairNet:} top-down semantic aggregation for accurate one shot
  detection.
\newblock In {\em WACV}, 2018.

\bibitem{kim2018san}
D.~K. Yonghyun~Kim, Bong-Nam~Kang.
\newblock San: Learning relationship between convolutional features for
  multi-scale object detection.
\newblock {\em ECCV}, 2018.

\bibitem{yu2018dln}
F.~Yu, D.~Wang, E.~Shelhamer, and T.~Darrell.
\newblock Deep layer aggregation.
\newblock In {\em CVPR}, 2018.

\bibitem{zhang2018refinedet}
S.~Zhang, L.~Wen, X.~Bian, Z.~Lei, and S.~Z. Li.
\newblock Single-shot refinement neural network for object detection.
\newblock In {\em CVPR}, 2018.

\bibitem{zhao2019m2det}
Q.~Zhao, T.~Sheng, Y.~Wang, Z.~Tang, Y.~Chen, L.~Cai, and H.~Ling.
\newblock M2det: A single-shot object detector based on multi-level feature
  pyramid network.
\newblock {\em AAAI}, 2019.

\bibitem{xu2018stdn}
P.~Zhou, B.~Ni, C.~Geng, J.~Hu, and Y.~Xu.
\newblock Scale-transferrable object detection.
\newblock In {\em CVPR}, 2018.

\bibitem{zoph2017nas}
B.~Zoph and Q.~V. Le.
\newblock Neural architecture search with reinforcement learning.
\newblock In {\em ICLR}, 2017.

\bibitem{zoph2017learning}
B.~Zoph, V.~Vasudevan, J.~Shlens, and Q.~V. Le.
\newblock Learning transferable architectures for scalable image recognition.
\newblock In {\em CVPR}, 2018.

\end{thebibliography}
}

\begin{appendices}

\section{Anytime Detection}
\label{sec:anytime_detection}
\begin{figure}[h]
    \centering
    \includegraphics[width=0.35\textwidth]{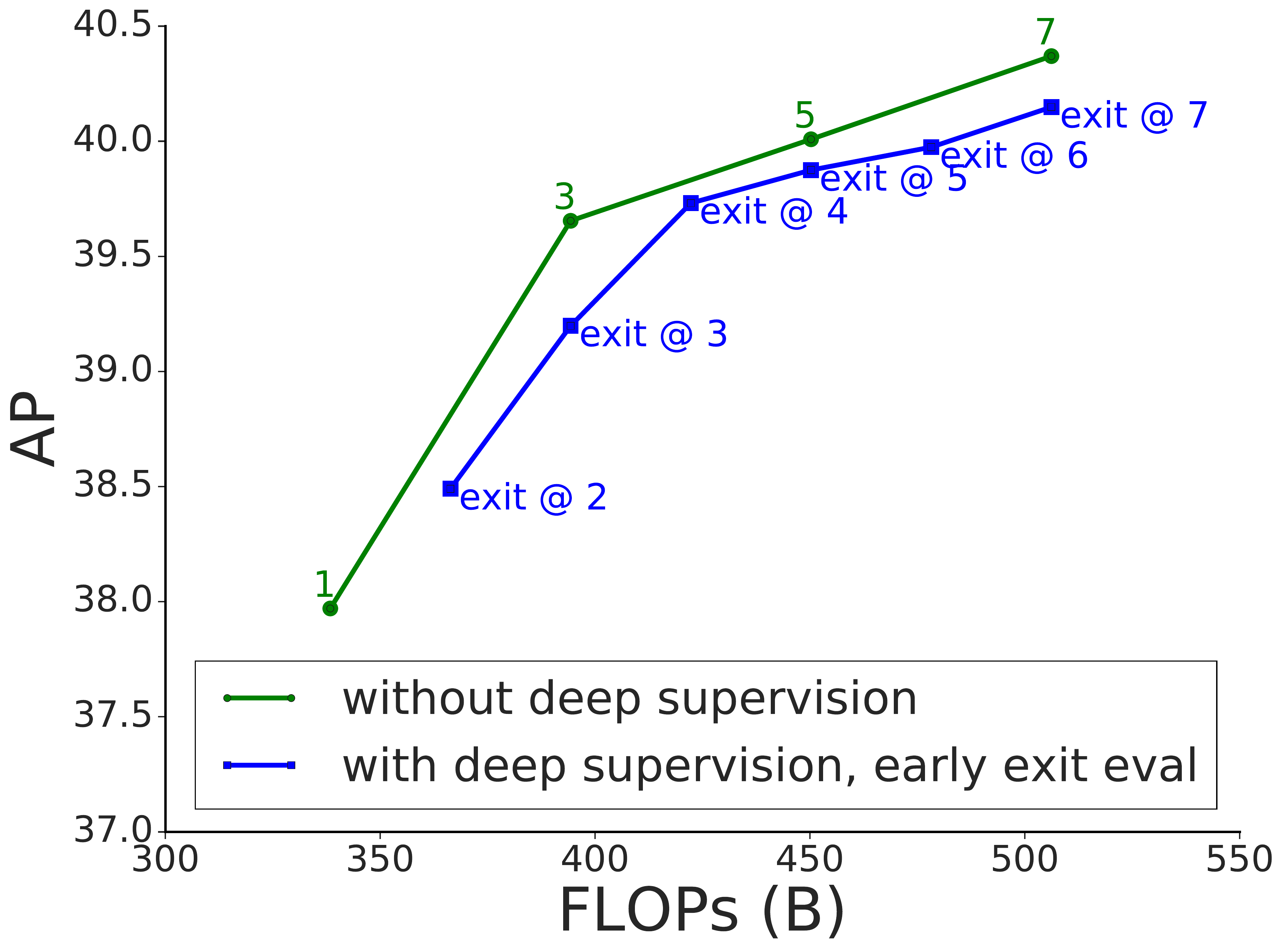} 
    \caption{Performance of model trained with and without deep supervision using R-50 @ 384 NAS-FPN. The model trained with deep supervision and tested with anytime detection has similar accuracy compared to model without deep supervision.}
    \label{fig:deepsv}
\end{figure}

The NAS-FPN architecture has the potential to output detections at any intermediate pyramid network. We designed the experiments to compare NAS-FPN models with and without anytime detection. We train and evaluate NAS-FPN with varying number of stacked pyramid layers as the baseline and compare the performances to model trained with deep supervision and generate detections with early exit. In Figure \ref{fig:deepsv}, the performance of anytime detection models is close to the baseline model, showing the evidence that NAS-FPN can be used for anytime detection.

\end{appendices}

\end{document}